\documentclass[sigconf]{acmart}
\AtBeginDocument{%
	}

\usepackage{booktabs}
\usepackage{algorithmic}
\usepackage[ruled,vlined]{algorithm2e}
\usepackage{multirow}

\begin{document}
	\title{Secure and Explainable Fraud Detection in Finance via Hierarchical Multi-source Dataset Distillation}
	
	\author{Yiming Qian}
	\email{Qian\_Yiming@a-star.edu.sg}
	\affiliation{%
		\institution{Institute of High Performance Computing, A*STAR}
		\country{Singapore}
	}
	
	\author{Thorsten Neumann}
	\email{thorsten.neumann@sc.com}
	\affiliation{%
		\institution{Standard Chartered Bank}
		\country{Singapore}
	}
	
	\author{Xueyining Huang}
	\email{xueyining.huang22@student.xjtlu.edu.cn}
	\affiliation{%
		\institution{Xi'an Jiaotong--Liverpool University}
		\country{China}
	}
	
	\author{David Hardoon}
	\email{DavidR.Hardoon@sc.com}
	\affiliation{%
		\institution{Standard Chartered Bank}
		\country{Singapore}
	}
	
	\author{Fei Gao}
	\email{gaofei@a-star.edu.sg}
	\affiliation{%
		\institution{Institute of High Performance Computing, A*STAR}
		\country{Singapore}
	}
	
	\author{Yong Liu}
	\email{liuyong@a-star.edu.sg}
	\affiliation{%
		\institution{Institute of High Performance Computing, A*STAR}
		\country{Singapore}
	}
	
	\author{Siow Mong Rick Goh}
	\authornote{Corresponding author.}
	\email{gohsm@a-star.edu.sg}
	\affiliation{%
		\institution{Institute of High Performance Computing, A*STAR}
		\country{Singapore}
	}
	
	\renewcommand{\shortauthors}{Qian et al.}
	
	\begin{abstract}
		We present an explainable, privacy-preserving dataset–distillation framework for collaborative fraud detection in finance. Our method converts a trained random forest into a set of transparent, axis-aligned rule regions (leaf hyperrectangles) and synthesizes transactions by uniformly sampling within each region. The resulting dataset is a compact, auditable surrogate of the original that preserves local feature interactions while avoiding direct exposure of sensitive records. Beyond privacy and utility, the rule regions provide explainable AI artifacts at both global and local levels: aggregated rule summaries (support, lift) reveal global structure, while per-case assignment to generating regions yields concise, human-readable rationales and calibrated uncertainty from tree-vote disagreement.
		
		Using the IEEE-CIS fraud dataset (590k transactions; three institution-style clusters), our distilled sets reduce volume by $85$–$93\%$ (typically $<15\%$ of the original size) while maintaining competitive performance: models trained on distilled data achieve comparable precision and micro-F1 with only a modest AUC reduction. Importantly, augmenting models with synthesized data across institutions substantially improves cross-cluster precision, recall, and AUC. Structural similarity between real and synthesized data exceeds $93\%$ via nearest-neighbor cosine analysis, while membership-inference attacks perform at chance when distinguishing real hold-out vs.\ training samples ($\mathrm{AUC}{\approx}0.50$), indicating low memorization risk. Filtering high-uncertainty synthetic points by disagreement scores further improves downstream AUC (up to $0.687$) and enhances calibration. Sensitivity analysis shows weak dependence on the distillation ratio (AUC $0.641{\to}0.645$ from $6\%$ to $60\%$).
		
		Overall, tree-region distillation offers a practical path to \emph{trustworthy} fraud analytics: interpretable global rules, per-case rationales with quantified uncertainty, and strong privacy guarantees—supporting real-time deployment and regulatory audit in multi-institution financial settings.
	\end{abstract}

	
	
	\maketitle
	
	\section{Introduction}
	Financial fraud—including payment card abuse, money laundering, synthetic identities, and account takeover—continues to rise in both volume and sophistication, costing the global economy tens of billions of dollars annually~\cite{nilson2024fraud}. In practice, suspicious activity is typically detected using fixed, rule-based systems established within individual institutions. While these rules are transparent, they lack adaptability. As institutions increasingly adopt machine learning models, predictive accuracy improves, but model opacity grows. This shift highlights a fundamental tension between performance and explainability: regulators and institutions demand decisions that are not only effective but also interpretable and auditable. Furthermore, fraud strategies that appear infrequently within a single institution’s data often require collaborative perspectives, yet such collaboration remains limited. Regulatory exposure risks (e.g., GDPR, PDPA, GLBA) and the difficulty of establishing compliant data-sharing agreements prevent the exchange of rules, models, or transaction data. As a result, most institutions operate in isolation, reducing recall of rare or newly emerging fraud patterns.
	
	A growing body of research attempts to reconcile collaboration, privacy, and transparency. Federated learning (FL) avoids raw data movement by exchanging parameter updates~\cite{goetz2020federated}, but vulnerabilities such as gradient inversion~\cite{huang2021evaluating} and membership inference~\cite{shokri2017membership} threaten sensitive information and hinder adoption. Cryptographic approaches like secure multi-party computation (SMPC)~\cite{evans2018pragmatic} and homomorphic encryption (HE)~\cite{acar2018survey} provide strong guarantees but incur prohibitive computational costs, limiting real-time deployment. Differential privacy (DP)~\cite{dwork2006differential} offers formal protections but often reduces minority-class recall—the very region most critical for fraud detection. From an explainability perspective, these methods further obscure decision pathways, offering little human-readable insight into why a transaction is flagged.
	
	Synthetic data generation has re-emerged as a complementary approach, enabling institutions to share representative but privacy-preserving records. Generative methods such as GANs~\cite{zhao2022synthesizing}, gradient matching~\cite{af3ad26036714fcbb4ffaf23ce6cee9f}, and diffusion models~\cite{moser2024latent} achieve high statistical fidelity but often sacrifice interpretability. Their complexity makes auditing difficult: determining whether a generated transaction memorizes an outlier, or how it maps to meaningful fraud typologies, is non-trivial. Deployment is further constrained by GPU requirements and operational overhead, creating barriers for widespread use in production systems.
	
	\textbf{Why tree-based hyperrectangles suit fraud data.}
	Fraudulent behavior often arises from sparse, segment-specific pockets of elevated risk that emerge from the \emph{co-occurrence of a few interpretable conditions} (e.g., new device \& cross-border \& atypical MCC \& off-hours). Random forests partition tabular features into axis-aligned leaves that align naturally with these pockets and with existing compliance rule engines. Treating each leaf as a hyperrectangle yields \emph{human-readable rule regions} (conjunctions of simple predicates) that: (i) respect mixed data types and heavy-tailed marginals without aggressive preprocessing; (ii) capture local interactions \emph{conditional on the decision path} rather than relying on a global generator; and (iii) provide direct operational artifacts for audit (support, lift, and explicit “if–then” rationales). Sampling \emph{within} these regions preserves the geometry of high-lift subpopulations while destroying pointwise identity, supporting privacy goals and calibrated uncertainty estimates via tree-vote dispersion. Computationally, random forest training and region-based resampling are CPU-efficient, enabling frequent refreshes to track adversarial drift without specialized hardware.
	
\begin{figure}[htbp]
		\centering
		\includegraphics[width=\columnwidth]{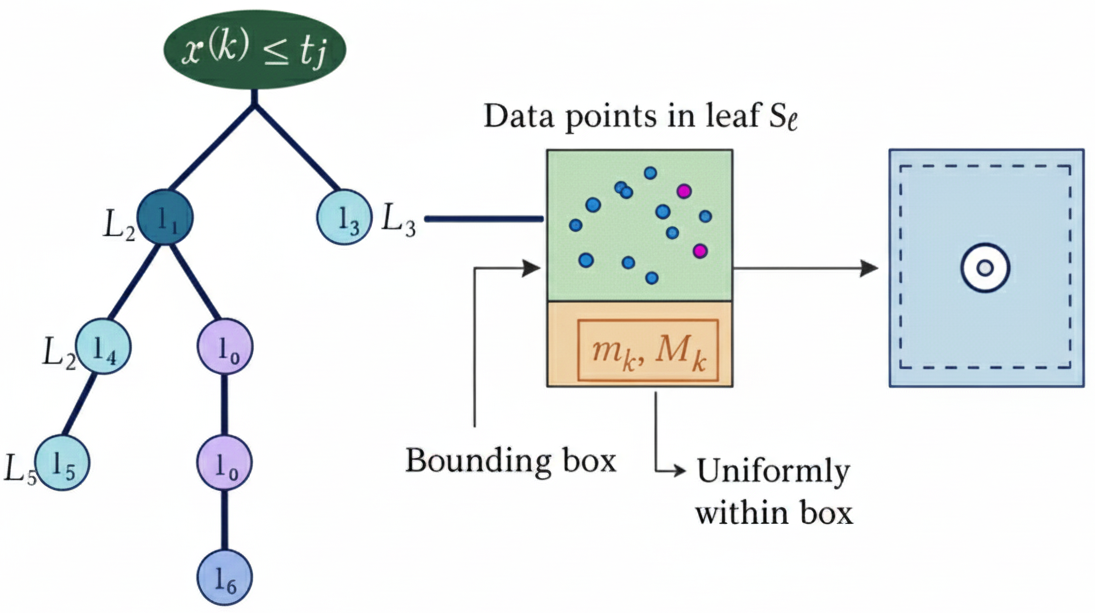}
		\caption{Illustration of tree-based dataset distillation.}
		\label{fig:algorithm_diagram}
	\end{figure}

	To address the combined needs of collaboration, privacy, and explainability, we propose a structure-aware dataset distillation approach that explicitly integrates these tree-derived rule regions. Our method (illustrated in Fig.~\ref{fig:algorithm_diagram}) leverages the partition geometry of random forest ensembles to extract axis-aligned hyperrectangles—effectively human-readable rule regions—that summarize coherent subpopulations of the data. Synthetic transactions are then generated by uniformly sampling within these regions, while optional disagreement-based filtering removes low-confidence samples. This framework provides multiple layers of transparency:  
	(i) globally, the distilled dataset is composed of explicit rule regions that can be aggregated and ranked by support and fraud lift;  
	(ii) locally, each synthetic sample inherits an interpretable predicate (“if–then” rule) describing its generation; and  
	(iii) uncertainty is quantified via random forest vote distributions, yielding calibrated measures of reliability.
	
	Unlike deep generative models, our approach is auditable, CPU-efficient, and directly aligned with regulatory requirements for interpretability in financial decision-making. It enables institutions not only to collaborate securely but also to justify and communicate decisions to auditors, regulators, and investigators.
	
	\begin{table*}[htbp]
		\centering
		\caption{Comparison of our method, diffusion, GAN, and federated learning.}
		\label{tab:compact_comparison}
		\renewcommand{\arraystretch}{1.2}
		\small
		\begin{tabular}{p{2.8cm} p{3.1cm} p{3.1cm} p{3.1cm} p{3.3cm}}
			\toprule
			\textbf{Aspect} & \textbf{Ours} & \textbf{Diffusion} & \textbf{GAN} & \textbf{Fed.~Learning} \\
			\midrule
			Shared artifact &
			Tree-region synthetic data (RF leaf sampling) &
			Synthetic samples (denoising) &
			Synthetic samples (generator) &
			Model updates (not reusable for future retraining) \\
			\addlinespace
			Transparency \& privacy &
			\textbf{High} explainability; low leakage (passes MIA) &
			Low–moderate; risk of memorization &
			Low–moderate; collapse/inversion risk &
			Limited explainability; gradient/MIA attacks possible \\
			\addlinespace
			Cost \& scalability &
			\textbf{Low} compute; fast resampling; CPU-friendly &
			High compute; slow sampling &
			Medium–high; unstable training &
			Medium–high; heavy infra and comm overhead \\
			\addlinespace
			Data coverage &
			Moderate; axis-aligned regions capture local structure &
			High; broad mode coverage with guidance &
			Medium; prone to collapse without stabilization &
			Depends on client heterogeneity; no reusable synthetic data \\
			\bottomrule
		\end{tabular}
	\end{table*}
	
	We validate our framework on the IEEE-CIS Fraud Detection dataset~\cite{ieee_fraud_detection_kaggle}, partitioned to emulate multiple institutions. Experiments demonstrate that our method reduces data volume by 85–93\% while retaining strong predictive performance, improves cross-institution recall when sharing distilled sets, and passes rigorous privacy evaluations such as membership inference attacks. Crucially, the explicit rule structures and calibrated uncertainties provide a pathway toward explainable, trustworthy fraud detection that balances performance, privacy, and regulatory accountability.
	
	Our contributions are as follows:
	\begin{enumerate}
		\item \textbf{Explainable, Privacy-Preserving Data Distillation:} A rule-region synthesis framework that generates transparent, auditable synthetic transactions for collaborative fraud detection.
		\item \textbf{Global and Local XAI Artifacts:} Rule summaries, per-sample rationales, and uncertainty estimates that enable both holistic model interpretation and transaction-level explanations.
		\item \textbf{Empirical Validation:} Experiments demonstrating strong performance retention, cross-institution gains, and resilience against memorization and inference attacks.
		\item \textbf{Practical Deployment:} A CPU-efficient, auditable approach that aligns with regulatory expectations for fairness, transparency, and accountability in financial services.
	\end{enumerate}
	
	\section{Related Work}
	Providing collaborative training for machine learning models is a long-standing research area with many proposed solutions. Two major approaches are federated learning and dataset distillation. Federated learning emphasizes sending model weights instead of sharing raw data, thereby enabling multi-source data training without direct data exchange. In contrast, dataset distillation focuses on creating a small synthetic dataset that can replace the original dataset for training new models. 
	
	\subsection{Dataset Distillation}
	There are many approaches designed to generate synthetic data from training samples. Meta-model matching methods optimize synthetic datasets via a bi-level process: a model is first (approximately) trained to convergence on the distilled data, and then the synthetic samples are iteratively updated to maximize performance on the original dataset. Representative methods include DD~\cite{wang2018dataset}, RFAD~\cite{loo2022efficient}, FRePO~\cite{zhou2022dataset}, and LinBa~\cite{deng2022remember}; however, the nested optimization makes these methods computationally demanding. Closely related—but more efficient—are gradient matching methods (DC~\cite{zhao2021dataset}, DCC~\cite{lee2022dataset}, IDC~\cite{kimICML22}), which replace the meta-objective with direct alignment of model gradients between synthetic and original data, substantially reducing optimization cost. Trajectory matching methods (MTT~\cite{cazenavette2022distillation}, TESLA~\cite{cui2023scaling}) extend this idea by aligning the entire sequence of parameter updates (training trajectories), capturing longer-term dynamics but reintroducing higher cost due to multi-step gradient unrolling. In contrast, distribution matching methods (DM~\cite{10030751}, CAFE~\cite{9879629}, SGDC~\cite{10.1145/3637528.3671682}) bypass meta-optimization by directly aligning data or feature distributions (e.g., via Maximum Mean Discrepancy), thereby improving scalability and speed. Distinct from the above, factorization-based methods (LinBa~\cite{deng2022remember}, HaBa~\cite{liu2022dataset}) structure the synthetic dataset into learnable bases and hallucinators to promote parameter sharing and reduce redundancy; factorization can be integrated into both meta-model and trajectory frameworks, offering a flexible trade-off between expressiveness and efficiency. Overall, these approaches share the goal of synthesizing compact, informative datasets, yet differ in their optimization strategies, computational burden, and the extent of training dynamics or distributional information they preserve.
	
	Generative dataset distillation replaces directly optimized synthetic instances with a learned generative prior whose latent space is tuned to yield a compact, high-utility surrogate dataset. In the GAN family, work progresses from trajectory-aligned latent priors (Deep Generative Prior)~\cite{cazenavette2023glad}, to distilling a compact generator as the distilled artifact (DiM)~\cite{wang2023dim}, to dual global–local objectives that preserve coarse structure while sharpening discriminative detail~\cite{li2024generative}. Diffusion-based approaches leverage stable denoising dynamics for broader mode coverage: minimax refinement for informative synthesis~\cite{gu2024efficient}, disentangled class/style factorization (D4M)~\cite{su2024d4m}, and influence-guided timestep weighting for representativeness~\cite{chen2025igd}. In summary, GAN priors emphasize fast sampling with structural regularization, whereas diffusion priors trade higher per-sample cost for improved mode coverage and principled guidance.
	
	\subsection{Federated and Collaborative Learning}
	Federated learning (FL) is a decentralized paradigm that enables multiple participants to collaboratively train a global model without exchanging their local raw data~\cite{pmlr-v54-mcmahan17a}. In the standard FL workflow, each client computes model updates (such as gradients or weights) on its local data, and a central server aggregates these updates to form a new global model. This iterative process allows model training across distributed datasets while preserving data locality and privacy. FL supports various system architectures, including cross-silo and cross-device federations, and incorporates techniques to address communication efficiency and update aggregation~\cite{9464278}. Despite its privacy advantages, FL introduces challenges such as significant communication overhead due to frequent model updates~\cite{9464278} and the impact of data and system heterogeneity on convergence and model quality~\cite{li2020federated}. Furthermore, even without sharing raw data, exchanged model parameters can be exploited for attacks such as membership inference and gradient inversion~\cite{7958568, huang2021evaluating}. To address these vulnerabilities, secure aggregation~\cite{10.1145/3133956.3133982} and differential privacy~\cite{dwork2006calibrating} have been incorporated, although these introduce additional computational and performance trade-offs.
	
	In contrast to these existing works, our method combines the interpretability and computational simplicity of tree-based data distillation with uncertainty-based filtering. This addresses critical privacy concerns highlighted in prior research while enabling practical deployment for collaborative fraud detection.
	
	\section{Dataset}
	Our experiments use the IEEE-CIS Fraud Detection dataset from a Kaggle competition~\cite{ieee_fraud_detection_kaggle}. After cleaning and removing rows and columns with missing values, the dataset contains 590{,}492 transactions with 29 features. We split the data into training and test sets with 472{,}393 and 118{,}099 samples, respectively. To emulate multiple institutions, we apply $k$-means clustering to partition the \emph{training} set into three clusters containing 138{,}944, 145{,}891, and 187{,}558 samples. These clusters serve as proxies for transactions from three independent financial institutions. Table~\ref{tab:stats} summarizes the dataset statistics.
	
	As shown in Table~\ref{tab:stats}, the dataset is highly imbalanced: positive (fraudulent) cases account for only 1.5–6.7\% within individual clusters and 3.5\% overall in the test set. This imbalance poses a well-known challenge in fraud detection: models optimized for overall accuracy can achieve high scores by predicting the majority class (non-fraud) while failing to capture rare but critical fraud cases. Consequently, precision, recall, and AUC are more informative evaluation metrics than accuracy, and special attention must be paid to thresholding and cost-sensitive learning.
	
	\begin{table}[htbp]
		\centering
		\caption{Statistics of the training data split into three clusters and the test set.}
		\label{tab:stats}
		\begin{tabular}{cccc}
			\toprule
			& Samples & Positives & Positive Ratio \\
			\midrule
			Cluster 1 & 138{,}944 & 3{,}885 & 2.8\% \\
			Cluster 2 & 145{,}891 & 9{,}795 & 6.7\% \\
			Cluster 3 & 187{,}558 & 2{,}844 & 1.5\% \\
			Test      & 118{,}099 & 4{,}131 & 3.5\% \\
			\bottomrule
		\end{tabular}
	\end{table}
	
\begin{figure}[htbp!]
		\centering
		\includegraphics[width=\columnwidth]{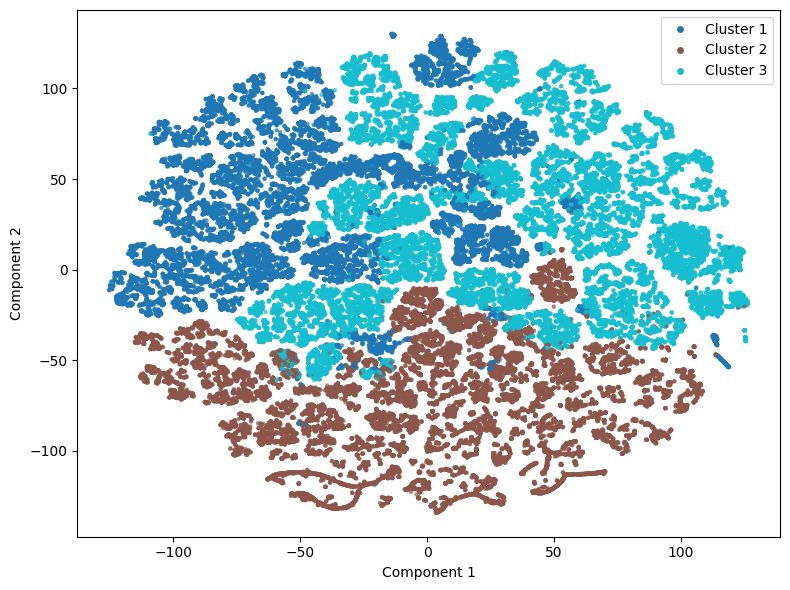}
		\caption{Visualization of the dataset partitioned into three clusters using $k$-means.}
		\label{fig:three_clusters}
	\end{figure}

	A t-SNE visualization~\cite{maaten2008visualizing} (Fig.~\ref{fig:three_clusters}) illustrates the three distinct clusters. The cleaned dataset will be made publicly available upon acceptance of the paper.
	
	\section{Method}
	We introduce a dataset distillation framework that leverages the partitioning structure of random forest classifiers to synthesize high-fidelity artificial data. The core idea is to exploit the local geometry captured by the leaf nodes of decision trees. Each leaf corresponds to an axis-aligned hyperrectangle that encloses a subset of samples with similar feature profiles, from which we can uniformly sample to generate synthetic points.
	
	Let $X = \{x_1, x_2, \dots, x_n\} \subset \mathbb{R}^d$
	denote the dataset with $d$ features. A trained random forest $F$ consists of an ensemble of decision trees. For a tree $T \in F$ and a sample $x_i$, denote the index of its leaf by
	\begin{equation}
		\ell_i = T.\mathrm{apply}(x_i).
	\end{equation}
	The set of samples in a leaf $\ell$ is defined as
	\begin{equation}
		S_\ell = \{\, x_j \in X \mid T.\mathrm{apply}(x_j) = \ell \,\}.
	\end{equation}
	
	For each feature dimension $k \in \{1,2,\dots,d\}$, compute the minimum and maximum values of the samples in $S_\ell$:
	\begin{equation}
		m_\ell^{(k)} = \min_{x \in S_\ell} x^{(k)}, 
		\quad 
		M_\ell^{(k)} = \max_{x \in S_\ell} x^{(k)}.
	\end{equation}
	This defines an axis-aligned bounding box (hyperrectangle):
	\begin{equation}
		H_\ell = \big\{\, x \in \mathbb{R}^d \;\big|\; m_\ell^{(k)} \leq x^{(k)} \leq M_\ell^{(k)},\; \forall k \,\big\}.
	\end{equation}
	
	A synthetic sample $\tilde{x}_\ell$ is generated by sampling each feature independently from a uniform distribution within the bounding box:
	\begin{equation}
		\tilde{x}_\ell^{(k)} \sim \mathcal{U}(m_\ell^{(k)}, \, M_\ell^{(k)}), 
		\quad \forall k.
	\end{equation}
	
	The final synthetic dataset is the union of such samples across all leaves of all trees in the forest:
	\begin{equation}
		\tilde{X} = \bigcup_{T \in F} \;\; \bigcup_{\ell \in \mathrm{Leaves}(T)} \tilde{x}_\ell.
	\end{equation}
	
	\begin{algorithm}[t]
		\caption{Dataset Distillation via Random Forest Leaf Sampling}
		\label{alg:rf_distillation}
		\KwIn{$F$ — trained RandomForestClassifier; 
			$X \in \mathbb{R}^{n \times d}$ — training dataset}
		\KwOut{$\tilde{X}$ — synthesized dataset}
		\BlankLine
		
		$\tilde{X} \gets \emptyset$\;
		\ForEach{tree $T \in F$}{
			$\text{LeafIndices} \gets T.\mathrm{apply}(X)$\;
			$\text{UniqueLeaves} \gets \mathrm{unique}(\text{LeafIndices})$\;
			
			\ForEach{leaf $\ell \in \text{UniqueLeaves}$}{
				$S_\ell \gets \{\, x_i \in X \mid \text{LeafIndices}[i] = \ell \,\}$\;
				
				\tcp{Compute bounding box across features}
				\For{$k \gets 1$ \KwTo $d$}{
					$m_k \gets \min_{x \in S_\ell} x^{(k)}$\;
					$M_k \gets \max_{x \in S_\ell} x^{(k)}$\;
				}
				
				\tcp{Sample uniformly within bounding box}
				$\tilde{x}_\ell \gets [\,]$\;
				\For{$k \gets 1$ \KwTo $d$}{
					$\tilde{x}_\ell^{(k)} \sim \mathcal{U}(m_k, M_k)$\;
				}
				
				$\tilde{X} \gets \tilde{X} \cup \{\tilde{x}_\ell\}$\;
			}
		}
		\Return{$\tilde{X}$}\;
	\end{algorithm}
	
	\section{Experiments}
	\subsection{Baseline Comparison}
	\paragraph{Diffusion baseline}
	As a generative comparator, we implement a conditional denoising diffusion model (DDPM)~\cite{su2024d} adapted to tabular fraud data. Features are standardized and noised forward using a linear or cosine variance schedule. An MLP denoiser with sinusoidal time embeddings predicts Gaussian noise conditioned on the class label, trained with the standard DDPM objective (mean-squared error between true and predicted noise). During synthesis, ancestral denoising steps generate synthetic transactions per class, drawing conditions from the empirical label prior. Optional DP--SGD training is explored to quantify the privacy–utility trade-off, though the main results use non-DP training for comparability. 
	Key parameters include $T{=}1000$ denoising steps, hidden dimensions [512, 512] in the MLP denoiser, GELU activations with layer normalization, Adam optimizer (learning rate $1\times10^{-4}$), batch size 512, and 200 training epochs. 
	
	\paragraph{GAN baseline}
	As another generative comparator, we implement a conditional Wasserstein GAN with gradient penalty (WGAN--GP)~\cite{gulrajani2017improved} for tabular synthesis. A multilayer perceptron (MLP) generator maps Gaussian noise vectors concatenated with class conditions into synthetic feature vectors, while a critic network jointly evaluates real and generated samples under the same conditions. Training follows the WGAN--GP objective with multiple critic updates per generator step and gradient penalty regularization to enforce Lipschitz continuity. During sampling, class conditions are drawn according to the empirical label distribution, and the generator produces synthetic transactions aligned with the observed class prior. 
	Key parameters include latent dimension $z{=}128$, two hidden layers of size 512 in both generator and critic, gradient penalty coefficient $\lambda{=}10$, $n_{\text{critic}}{=}5$ critic updates per generator step, Adam optimizers with $(\beta_1{=}0, \beta_2{=}0.99)$, and batch size 512 trained for 200 epochs.
	
	\subsection{Experimental Results}
	We simulate a scenario involving three financial institutions, each with its own transaction data but unable to share flagged suspicious transactions with other organizations for collaborative model training. Three clusters of data represent these unaffiliated institutions. Each institution trained a random forest classifier to identify fraudulent transactions. We performed three independent distillation processes on the three real data clusters, consistently achieving a $\sim$90\% data reduction rate.
	
	\begin{table}[htbp]
		\caption{Statistics of Synthesized Datasets.}
		\centering
		\begin{tabular}{cccc}
			\toprule
			\textbf{Data Source} & \textbf{Real}    & \textbf{Synthesized} & \textbf{Ratio}  \\  \midrule
			Cluster 1 & 138,944 & 13,812    & 9.9\%  \\
			Cluster 2 & 145,891 & 19,927    & 13.7\% \\
			Cluster 3 & 187,558 & 13,768    & 7.3\% \\ 
			\bottomrule
		\end{tabular}
	\end{table}
	
	We measured precision, recall, micro-F1, and AUC across real and synthesized datasets. The synthetic data generated by our algorithm are indicated in brackets. Our results show a slight improvement in precision when training on synthesized data, while micro-F1 remains comparable and AUC decreases marginally. As expected, AUC—which reflects a model’s ability to discriminate between classes—is somewhat lower on synthesized datasets.  
	
	We also evaluated the scenario of combining all real data versus combining all synthesized data to train a classifier. As anticipated, we observed a $\sim$20\% reduction in AUC when training on the combined synthesized dataset compared to the combined real dataset.
	
	\begin{table}[htbp!]
		\centering
		\caption{Performance Metrics Across Groups and Synthesized Variants.}
		\begin{tabular}{@{}lcccc@{}}
			\toprule
			\textbf{Data Type}       & \textbf{Precision} & \textbf{Recall} & \textbf{Micro-F1} & \textbf{AUC} \\ \midrule
			Cluster 1             & 0.900 & 0.144 & 0.969 & 0.567 \\
			Cluster 1 (Synth.)    & 0.909 & 0.012 & 0.965 & 0.505 \\
			Cluster 2             & 0.843 & 0.435 & 0.977 & 0.716 \\
			Cluster 2 (Synth.)    & 0.884 & 0.273 & 0.973 & 0.636 \\
			Cluster 3             & 1.000 & 0.093 & 0.968 & 0.547 \\
			Cluster 3 (Synth.)    & 0.948 & 0.010 & 0.965 & 0.505 \\ 
			Combined Real Data    & 0.943 & 0.604 & 0.985 & 0.801 \\ 
			\bottomrule
		\end{tabular}
	\end{table}
	
	The evaluation was conducted under a highly imbalanced test set, with a positive rate of only 3.5\%. In such cases, conventional accuracy or micro-averaged scores can be misleading, since a trivial classifier predicting all samples as negative would already achieve $\sim$96.5\% accuracy. The random baseline reflects this imbalance: while recall is 0.494, its precision (0.035) essentially matches the base rate, indicating near-random guessing.  
	
	Our method achieves much higher precision (0.910), meaning positive predictions are overwhelmingly correct. However, recall (0.287) remains modest, with fewer than one in three positives identified. Generative baselines (WGAN-GP and DDPM) trade precision for recall, yielding slightly higher AUC values as they capture more positives at the cost of increased false alarms. These results illustrate the precision–recall trade-off under severe imbalance and emphasize the need for threshold tuning and cost-sensitive evaluation in real-world fraud detection, where the relative costs of false positives and false negatives differ significantly.
	
	\begin{table}[htbp!]
		\centering
		\caption{Performance Metrics Between State-of-the-Art Algorithms.}
		\begin{tabular}{@{}lcccc@{}}
			\toprule
			\textbf{Data Type} & \textbf{Precision} & \textbf{Recall} & \textbf{Micro-F1} & \textbf{AUC} \\ \midrule
			Random Data        & 0.035 & 0.494 & 0.502 & 0.498 \\ 
			Ours Combined      & \textbf{0.910} & \textbf{0.287} & \textbf{0.974} & 0.643 \\
			WGAN-GP            & 0.753 & 0.217 & 0.336 & 0.795 \\
			DDPM               & 0.787 & 0.286 & 0.419 & \textbf{0.824} \\
			\bottomrule
		\end{tabular}
	\end{table}
	
	\subsection{Cross-Cluster Evaluation}
	We conducted a cross-cluster evaluation to study the impact of training a model on one cluster and testing its performance on another cluster. The experiments were carried out in two parts. First, we trained a random forest classifier on one cluster and tested it on a different cluster. Next, we repeated the same process, but augmented the training cluster with the aggregated synthetic data from all clusters. The results are presented in Table~\ref{tab:cross_cluster_performance}. With synthesized dataset augmentation, there is a significant performance gain in the cross-cluster evaluation.
	
	\begin{table}[htbp!]
		\centering
		\caption{Cross-cluster classification performance. Each row reports the results of training on one cluster and testing on another. ``+ Syn.'' denotes that the training cluster was augmented with aggregated synthetic data from Clusters 1, 2, and 3.}
		\label{tab:cross_cluster_performance}
		\begin{tabular}{cccccc}
			\toprule
			\textbf{Train} & \textbf{Test} & \textbf{Precision} & \textbf{Recall} & \textbf{AUC} \\
			\midrule
			Cluster 1 & Cluster 2 & 0.672 & 0.026 & 0.512 \\
			Cluster 1 & Cluster 3 & 0.927 & 0.103 & 0.551 \\
			Cluster 2 & Cluster 1 & 0.225 & 0.055 & 0.525 \\
			Cluster 2 & Cluster 3 & 0.605 & 0.082 & 0.541 \\
			Cluster 3 & Cluster 1 & 0.993 & 0.076 & 0.538 \\
			Cluster 3 & Cluster 2 & 0.988 & 0.008 & 0.504 \\ \midrule
			Cluster 1 + Syn.&Cluster 2& 0.896 & 0.433 & 0.715 \\
			Cluster 1 + Syn.&Cluster 3& 0.949 & 0.098 & 0.549 \\
			Cluster 2 + Syn.&Cluster 1& 0.998 & 0.988 & 0.994 \\
			Cluster 2 + Syn.&Cluster 3& 0.975 & 0.083 & 0.541 \\
			Cluster 3 + Syn.&Cluster 1& 0.998 & 0.989 & 0.994  \\
			Cluster 3 + Syn.&Cluster 2& 0.899 & 0.439 & 0.718  \\
			\bottomrule
		\end{tabular}
	\end{table}
	
	\subsection{Classifier Comparison}
	To evaluate whether synthesized datasets produced by our distillation pipeline generalize across diverse model families (beyond the random forest used for generation), we trained a suite of standard fraud detection classifiers on (i) the combined real dataset and (ii) the combined synthetic data. Table~\ref{tab:classifier_comparison} reports test AUC for each model. Across eleven classifier families, linear and generative models (Logistic Regression, Naive Bayes, QDA, AdaBoost) exhibit AUC improvements of 0.045–0.107 when trained on distilled data, while complex tree and instance-based models show 0.058–0.196 absolute declines, indicating that distillation smooths local boundary detail yet regularizes noise in favor of simpler learners.
	
	\begin{table}[htbp]
		\centering
		\caption{Test AUC of diverse classifiers trained on combined real vs. combined synthetic data. Boldface indicates (within each row) the higher AUC between real and synthesized training.}
		\label{tab:classifier_comparison}
		\begin{tabular}{lcc}
			\toprule
			\textbf{Classifier} & \textbf{Real} & \textbf{Synthesized} \\
			\midrule
			Random Forest & \textbf{0.801} & 0.643 \\
			Logistic Regression & 0.520 & \textbf{0.565} \\
			MLP & \textbf{0.654} & 0.639 \\
			XGBoost & \textbf{0.727} & 0.649 \\
			LSTM & \textbf{0.692} & 0.644 \\
			SVM (RBF) & \textbf{0.545} & 0.521 \\
			KNN & \textbf{0.718} & 0.585 \\
			Decision Tree & \textbf{0.831} & 0.635 \\
			Naive Bayes & 0.519 & \textbf{0.626} \\
			Quadratic Discriminant Analysis & 0.569 & \textbf{0.653} \\
			AdaBoost & 0.546 & \textbf{0.635} \\
			\bottomrule
		\end{tabular}
	\end{table}
	
	\subsection{Uncertainty Analysis for Data Synthesis}
	We follow Mentch et al.~\cite{mentch2016quantifying} to compute a disagreement score from the random forest classifier. The $j$-th input sample $\mathbf{x}_j$ is passed through the forest to obtain predictions from all trees, and the proportion of trees that vote for each class is computed. The disagreement score $d_j$ is defined as one minus the largest vote proportion:
	\begin{equation}
		d_j
		= 1 - \max_{c}\,\frac{1}{T}\sum_{t=1}^T \mathbb{I}\{f_t(\mathbf{x}_j)=c\},
		\quad j=1,\dots,M,
	\end{equation}
	where $T$ denotes the total number of trees, $C$ is the number of classes, $c \in \{1,\dots,C\}$ indexes classes, $f_t(\mathbf{x}_j) \in \{1,\dots,C\}$ is the $t$-th tree’s prediction, and $\mathbb{I}\{\cdot\}$ denotes the indicator function.
	
\begin{figure}[htbp]
		\centering
		\includegraphics[width=\columnwidth]{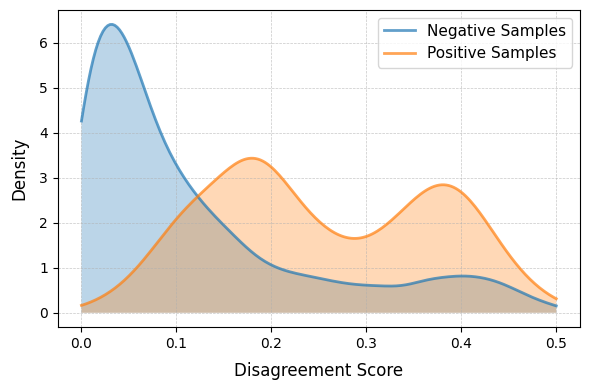}
		\caption{Disagreement score distribution of synthesized positive and negative samples.}
		\label{fig:positive_vs_negative}
	\end{figure}

	We computed the disagreement scores for our synthetic samples, and the score distributions for positive and negative samples are shown in Fig.~\ref{fig:positive_vs_negative}. The disagreement scores for negative samples are concentrated in the low-score region, whereas positive samples exhibit peaks around 0.18 and 0.37, suggesting greater uncertainty among positives. By setting an appropriate threshold, we can filter out samples with higher uncertainty, thereby improving overall data quality.
	
\begin{figure}[htbp]
		\centering
		\includegraphics[width=\columnwidth]{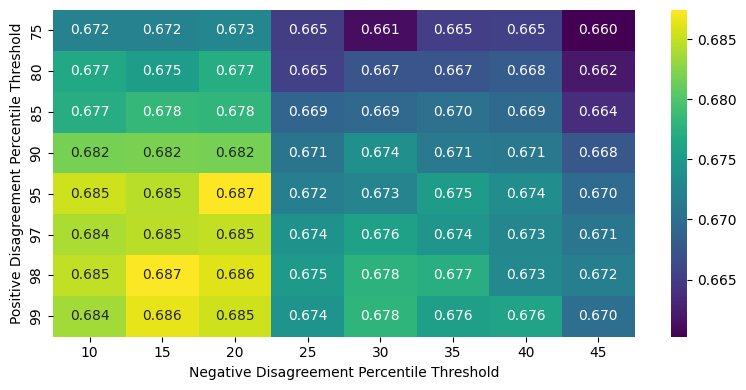}
		\caption{Grid search over positive and negative disagreement percentile thresholds. Cell values denote downstream AUC.}
		\label{fig:grid_search}
	\end{figure}

	We conducted an ablation to assess the effect of applying separate disagreement-score thresholds to filter synthesized positive and negative samples. A grid search over percentile cutoffs identified an optimal pair. The results (Fig.~\ref{fig:grid_search}) show that using the 95th percentile threshold for positives and the 20th percentile threshold for negatives yields the best performance, achieving an AUC of 0.687.
	
	\subsection{Sensitivity Analysis on Distillation Ratio}
	To evaluate sensitivity to the distillation ratio, we varied the quantity of synthetic samples by adjusting the number of trees in the random forest generator. Raising the ratio from 6\% to 60\% resulted in only a marginal improvement in test-set AUC (0.641 to 0.645), suggesting that downstream performance is largely insensitive to moderate increases in synthesized sample density.
	
	\begin{figure}[htbp]
		\centering
		\includegraphics[width=7cm]{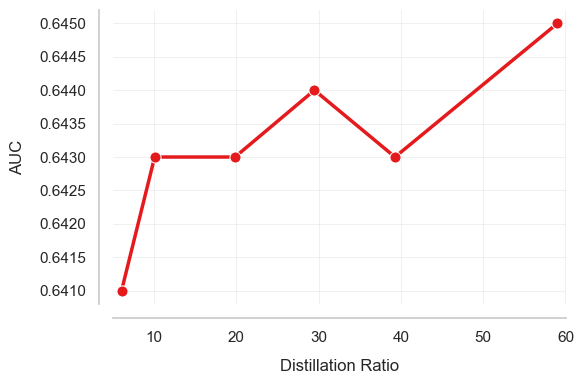}
		\caption{Test-set AUC as a function of the training-data distillation ratio.}
		\label{fig:distillation_ratio}
	\end{figure}
	
	\section{Memorization Risk Assessment}
	Synthetic data are algorithmically generated from the original data and may be subject to memorization risk. We employ three measures to assess the extent of memorization. First, we use t-SNE to visually inspect distributional overlap between synthesized and original data. Next, we calculate nearest-neighbor cosine similarity to quantify similarity between the synthesized and original datasets, helping balance fidelity and privacy. If the synthesized dataset is too dissimilar, it may fail to capture the characteristics of the original dataset; if it is too similar, it may introduce memorization and privacy risks. Third, we conduct a membership inference attack (MIA)~\cite{shokri2017membership} to check whether the synthesized data contain memorized records from the original dataset.
	
	\subsection{Visualization}
\begin{figure}[htbp]
		\centering
		\includegraphics[width=\columnwidth]{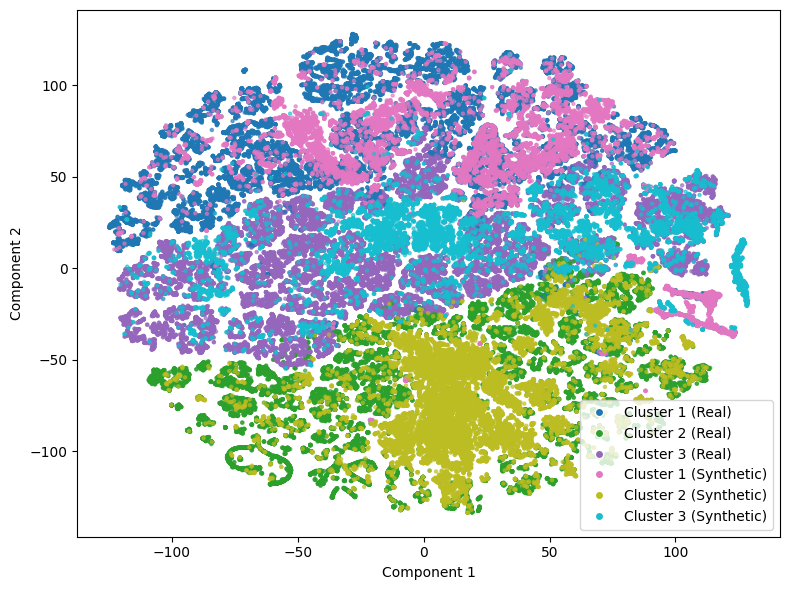}
		\caption{t-SNE visualization of real (shaded) and synthesized (dots) data across three clusters.}
		\label{fig:three_clusters_syn}
	\end{figure}

	We used t-SNE to visualize the three clusters of training data and their respective synthetic data (Fig.~\ref{fig:three_clusters_syn}). The figure displays three distinct clusters, with the synthetic data closely overlapping the real data. To quantify similarity between real and synthesized data, we compute nearest-neighbor cosine similarity.
	
	\subsection{Nearest-Neighbor Cosine Similarity}
	To quantify local similarity, we compute, for each synthetic vector $\mathbf{s}_i$, its maximum cosine similarity to any real vector $\mathbf{o}_j$.
	
	We conduct two comparisons: (i) similarity between the original and the synthesized dataset, and (ii) similarity between the original data and other clusters (as a baseline). The similarity metric is based on nearest-neighbor cosine similarity between source and target datasets.
	
	Formally, let the synthesized dataset be \( S = \{\mathbf{s}_1, \ldots, \mathbf{s}_N\} \) and the original dataset be \( O = \{\mathbf{o}_1, \ldots, \mathbf{o}_M\} \). For each synthetic vector \( \mathbf{s}_i \), define its similarity to the original set as the maximum cosine similarity to any real vector:
	\begin{equation}
		\max_{j} \; \mathrm{cos\_sim}(\mathbf{s}_i, \mathbf{o}_j) 
		= \max_{j} \left( \frac{\mathbf{s}_i \cdot \mathbf{o}_j}{\|\mathbf{s}_i\| \|\mathbf{o}_j\|} \right).
	\end{equation}
	The overall similarity between datasets is then the average of these nearest-neighbor scores:
	\begin{equation}
		\mathrm{Similarity}(S, O) 
		= \frac{1}{N} \sum_{i=1}^{N} \max_{j} \; \mathrm{cos\_sim}(\mathbf{s}_i, \mathbf{o}_j).
	\end{equation}
	
	A score close to 1 indicates that every synthetic sample closely aligns (in cosine space) with at least one real sample, reflecting strong geometric similarity. In our results, the synthetic data achieved more than 93\% similarity, whereas cross-cluster comparisons yielded only 63\%–73\% similarity.
	
	\begin{figure}[htbp!]
		\centering
		\includegraphics[width=7.5cm]{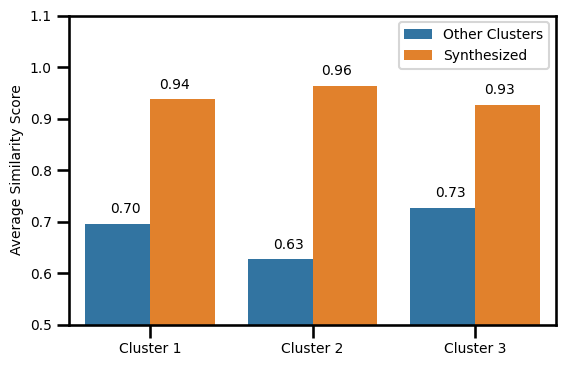}
		\caption{Nearest-neighbor cosine similarity between real clusters and synthesized datasets.}
		\label{fig:synthesized_vs_real}
	\end{figure}
	
	\subsection{Membership Inference Attack}
	Nearest-neighbor cosine similarity analysis showed that our synthesized dataset shares 93\% similarity with the original data. We next evaluate memorization and privacy risks using a membership inference attack (MIA)~\cite{shokri2017membership}. A shadow random forest model was first trained on half of the real training data. A logistic regression attack model was then fitted on the shadow model’s prediction probabilities, with the goal of distinguishing between “member” samples (from the shadow model’s training data) and “non-member” samples (from the held-out data).
	
	To assess memorization, we applied the attack model to predictions of a target random forest trained on the full real training dataset. Evaluation was performed across three groups: (1) real training samples (“members”), (2) real hold-out test samples (“non-members”), and (3) synthetic samples.
	
	The attack achieved an AUC of 0.502 when distinguishing real training samples from real hold-out samples, consistent with random guessing. This demonstrates that the attack model cannot reliably identify which real samples were used in training. However, when distinguishing real training samples from synthetic data, the attack AUC increased to 0.875.
	
	This elevated AUC does not indicate memorization; rather, it shows that the synthesized datasets are systematically different from the real training data—so different that the attack can easily tell them apart. Such separability is desirable from a privacy perspective: if synthesized records were simple copies or near-duplicates of real training points, the AUC would be closer to 0.5, as the attack model would be unable to distinguish them from true members. Instead, the result confirms that the synthetic generation method produces samples distinct from the training set, mitigating risks of direct memorization and overfitting.
	
	Moreover, the mean predicted membership probability assigned to synthetic samples was 0.5, further reinforcing that the attack model has no particular tendency to classify synthetic points as likely training members.
	
	In summary, the combination of (i) a high attack AUC between real and synthetic datasets, (ii) a mean membership probability of 0.5 for synthetic samples, and (iii) random-guess performance on real hold-out data confirms that our synthetic data generation approach avoids memorization of training records and provides strong privacy protection against membership inference attacks.
	
	\section{Conclusion}
	We introduced a tree-region dataset–distillation framework that turns random-forest leaves into transparent rule regions and uniformly samples within them to generate privacy-preserving synthetic transactions. This simple, CPU-friendly mechanism produces explicit, auditable “if–then” predicates for every synthesized record, enabling global rule summaries and per-case rationales with calibrated uncertainty from vote dispersion. Across a three-institution setting built from the IEEE-CIS dataset, the approach reduces data volume by 85–93\% while retaining strong downstream utility, improves cross-institution generalization when distilled sets are shared, and withstands membership-inference evaluation (near-chance on real hold-out vs.\ train), indicating low memorization risk. The explicit rule geometry also aligns with compliance workflows, easing model review and regulatory audits.
	
	Beyond serving as an operationally practical alternative to heavier generative pipelines, the method provides a trustworthy path to collaboration: institutions can exchange interpretable, reusable synthetic corpora instead of raw data or opaque model updates. Looking forward, we see four promising extensions: (i) privacy tightening via DP noise on region boundaries and count-based sampling; (ii) richer handling of mixed/categorical features and temporal sequences; (iii) adaptive, cost-sensitive thresholding to address severe class imbalance; and (iv) governance tooling for multi-party curation, drift monitoring, and audit trails. Taken together, these directions can turn tree-region distillation into a turnkey primitive for secure, explainable analytics at scale in regulated financial environments.
	
	\bibliographystyle{ACM-Reference-Format}
	\bibliography{sample-base}
\end{document}